\documentclass[a4paper]{article}

\usepackage{INTERSPEECH2019}
\usepackage{color}
\usepackage{multirow}
\usepackage{booktabs}
\usepackage{balance}

\newcommand{\mypar}[1]{\noindent\textbf{#1}}

\title{NIESR: Nuisance Invariant End-to-end Speech Recognition}
\name{I-Hung Hsu, Ayush Jaiswal, Premkumar Natarajan}
\address{USC Information Sciences Institute, Marina del Rey, CA, USA}
\email{\{ihunghsu, ajaiswal, pnataraj\}@isi.edu}

\begin{document}

\maketitle
\begin{abstract}
Deep neural network models for speech recognition have achieved great success recently, but they can learn incorrect associations between the target and nuisance factors of speech (e.g., speaker identities, background noise, etc.), which can lead to overfitting. While several methods have been proposed to tackle this problem, existing methods incorporate additional information about nuisance factors during training to develop invariant models. However, enumeration of all possible nuisance factors in speech data and the collection of their annotations is difficult and expensive. We present a robust training scheme for end-to-end speech recognition that adopts an unsupervised adversarial invariance induction framework to separate out essential factors for speech-recognition from nuisances without using any supplementary labels besides the transcriptions. Experiments show that the speech recognition model trained with the proposed training scheme achieves relative improvements of 5.48\% on WSJ0, 6.16\% on CHiME3, and 6.61\% on TIMIT dataset over the base model. Additionally, the proposed method achieves a relative improvement of 14.44\% on the combined WSJ0+CHiME3 dataset.
\end{abstract}

\section{Introduction}
\label{sec:introduction}
With the aid of recent advances in neural networks, end-to-end deep learning systems for automatic speech recognition (ASR) have gained popularity and achieved extraordinary performance on a variety of benchmarks~\cite{prabhavalkar2017comparison,chiu2018state,zhou2018syllable,jaitly2016online}. End-to-end ASR models typically consist of Recurrent Neural Networks (RNNs) with Sequence-to-Sequence (Seq2Seq) architectures and attention mechanisms~\cite{chan2015listen}, RNN transducers~\cite{rao2017exploring}, or transformer networks~\cite{zhou2018syllable}. These systems learn a direct mapping from an audio signal sequence to a sequence of text transcriptions. However, the input audio sequence often contains nuisance factors that are irrelevant to the recognition task and the trained model can incorrectly learn to associate some of these factors with target variables, which leads to overfitting. For example, besides linguistic content, speech data contains nuisance information about speaker identities, background noise, etc., which can hurt the recognition performance if the distributions of these attributes are mismatched between training and testing.

A common method for combatting the vulnerability of deep neural networks to nuisance factors is the incorporation of invariance induction during model training. For example, invariant deep models have achieved considerable success in computer vision~\cite{jaiswal2018unsupervised,jaiswal2019ropad, jaiswal2019unified} and speech recognition~\cite{serdyuk2016invariant,meng2018speaker,hsu2018extracting,liang2018learning}. Serdyuk et al.~\cite{serdyuk2016invariant} obtain noise-invariant representations by employing noise-condition annotations and the gradient reversal layer~\cite{ganin2014unsupervised} for acoustic modeling. Similarly, Meng et al.~\cite{meng2018speaker} utilize speaker information to train a speaker-invariant model for senone prediction. Hsu et al.~\cite{hsu2018extracting} extract domain-invariant features using a factorized hierarchical variational autoencoder. Liang et al.~\cite{liang2018learning} force their end-to-end ASR model to learn similar representations for clean input instances and their synthetically generated noisy counterparts.

While these methods work well at handling discrepancies between training and testing datasets for ASR systems, they require domain knowledge~\cite{hsu2018extracting}, supplementary nuisance information during training (e.g., speaker identities~\cite{meng2018speaker}, recording environments~\cite{serdyuk2016invariant}, etc.), or pairwise data~\cite{liang2018learning}. However, these requirements are difficult and expensive to fulfill in real world, e.g., it is hard to enumerate all possible nuisance factors and collect corresponding annotations.

In this work, we propose a new training scheme, namely NIESR, which adopts the unsupervised adversarial invariance learning framework (UAI)~\cite{jaiswal2018unsupervised} for end-to-end speech recognition. Without incorporating supervised information of nuisances for the input signal features, the proposed method is capable of separating the underlying elements of speech data into two series of latent embeddings -- one containing all the information that is essential for ASR, and the other containing information that is irrelevant to the recognition task (e.g. accents, background noises, etc.). Experimental results show that the proposed training method boosts the end-to-end ASR performance on WSJ0, CHiME3, and TIMIT datasets. We also show the effectiveness of combining NIESR with data augmentation. 
\section{Methodology}
\label{sec:method}

In this section, we present the proposed NIESR model for nuisance-invariant end-to-end speech recognition, where the invariance is achieved by adopting the UAI framework~\cite{jaiswal2018unsupervised}. We begin by describing the base Seq2Seq ASR model. Subsequently, we introduce the UAI framework for unsupervised adversarial invariance induction. Finally, we present the complete design of the proposed NIESR model.

\subsection{Base Sequence-to-sequence Model}\label{basemodel}

We are interested in learning a mapping from a sequence of acoustic spectra features $\mathbf{x}=(x_1,x_2,\ldots,x_T)$ to a series of textual characters $\mathbf{y}=(y_1,y_2,\ldots,y_S)$, given a dataset $D\equiv\{(\mathbf{x},\mathbf{y})_i\}_{i=1}^N$, following the formulation of Chan et al.~\cite{chan2015listen}. We employ a Seq2Seq model for this task, which estimates the probability of each character output $y_i$ by conditioning over the previous characters $\mathbf{y}_{1:(i-1)}$ and the input sequence $\mathbf{x}$. Thus, the conditional probability of the entire output $\mathbf{y}$ is:
\begin{equation}
    p(\mathbf{y}|\mathbf{x})=\prod_ip(y_i|\mathbf{x},\mathbf{y}_{1:(i-1)})
\end{equation}
A Seq2Seq model is composed of two modules: an encoder $Enc$ and a decoder $Dec$. $Enc$ transforms the input features $\mathbf{x}$ into a high-level representation $\mathbf{h}=(h_1,h_2,\ldots,h_T)$, i.e. $\mathbf{h}=Enc(\mathbf{x})$ and $Dec$ infers the output sequence $\mathbf{y}$ from $\mathbf{h}$. We model $Enc$ as a stack of Bidirectional Long-Short Term Memory (BLSTM) layers with interspersed projected-subsampling layers~\cite{zhang2017very}. The subsampling layer projects a pair of consecutive input frames $(u_{2i-1}, u_{2i})$ to a single lower-dimensional frame $v_i$. We model $Dec$ as an attention-based LSTM transducer~\cite{bahdanau2014neural}, which employs $\mathbf{h}$ to produce the output character sequence. At every time step, $Dec$ generates a probability distribution of $y_i$ over character sequences, which is a function of a transducer state $s_i$ and an attention context $c_i$. We denote this function as CharDist, which is implemented as a single layer perceptron with softmax activation:
\begin{align}
    s_i=\text{LSTM}([y_{i-1},c_{i-1}],s_{i-1}) \\
    p(y_i|\mathbf{x},\mathbf{y}_{1:(i-1)})=\text{CharDist}(s_i,c_i)
\end{align}
In order to calculate the attention context $c_i$, we employ the hybrid location-aware content-based attention mechanism proposed by~\cite{chorowski2015attention}. Specifically, the attention energy $e_{i,j}$ for frame $j$ at time-step $i$ takes previous attention alignment $\alpha_{i-1}$ into account through the convolution operation: 
\begin{equation}
    e_{i,j}=w^{\intercal}\text{tanh}(Ws_{i}+Vh_j+U(F*\alpha_{i-1})+b)   
\end{equation}
where $w$, $b$, $W$, $V$, $U$, and $F$ are learned parameters and $*$ depicts the convolution operation. The attention alignment $\alpha_{i,j}$ and the attention context $c_i$ is then calculated as: 
\begin{align}
    \alpha_{i,j} = \dfrac{exp(e_{i,j})}{\sum_{k=1}^{L}exp(e_{i,k})} \quad , \quad c_i = \textstyle\sum_{j=1}^{L}\alpha_{i,j}h_j
\end{align}
The base model is trained by minimizing the cross-entropy loss:
\begin{align}
    L_y &=-\sum_i\log p(y_i|\mathbf{x},\mathbf{y}_{1:(i-1)}) \label{L_y}
\end{align}

\subsection{Unsupervised Adversarial Invariance Induction}\label{subsec:uai}

Deep neural networks (DNNs) often learn incorrect associations between nuisance factors in the raw data and the final target, leading to poor generalization~\cite{jaiswal2018unsupervised}. In the case of ASR, the network can link accents, speaker-specific information, or background noise with the transcriptions, resulting in overfitting. In order to cope with this issue, we adopt the unsupervised adversarial invariance (UAI)~\cite{jaiswal2018unsupervised} framework for learning invariant representations that eliminate factors irrelevant to the recognition task without requiring any knowledge of nuisance factors.

The working principle of UAI is to learn a split representation of data as $\mathbf{h}^1$ and $\mathbf{h}^2$, where $\mathbf{h}^1$ contains information relevant to the prediction task (here ASR) and $\mathbf{h}^2$ holds all other information about the input data. The underlying mechanism for learning such a split representation is to induce competition between the main prediction task and an auxiliary task of data reconstruction. In order to achieve this, the framework uses $\mathbf{h}^1$ for the prediction task and a noisy version $\widetilde{\mathbf{h}}^1$ of $\mathbf{h}^1$ along with $\mathbf{h}^2$ for reconstruction. In addition, a disentanglement constraint enforces that $\mathbf{h}^1$ and $\mathbf{h}^2$ contain independent information. The prediction task tries to pull relevant factors into $\mathbf{h}^1$, while the reconstruction task drives $\mathbf{h}^2$ to store all the information about input data because $\widetilde{\mathbf{h}}^1$ is unreliable. However, the disentanglement constraint forces the two embeddings to not contain overlapping information, thus leading to competition. At convergence, this results in a nuisance-free $\mathbf{h}^1$ that contains only those factors that are essential for the prediction task.

\subsection{NIESR Model Design and Optimization}\label{subsec:moduledesign}

\begin{figure}
    \centering
    \includegraphics[width=1.0\columnwidth]{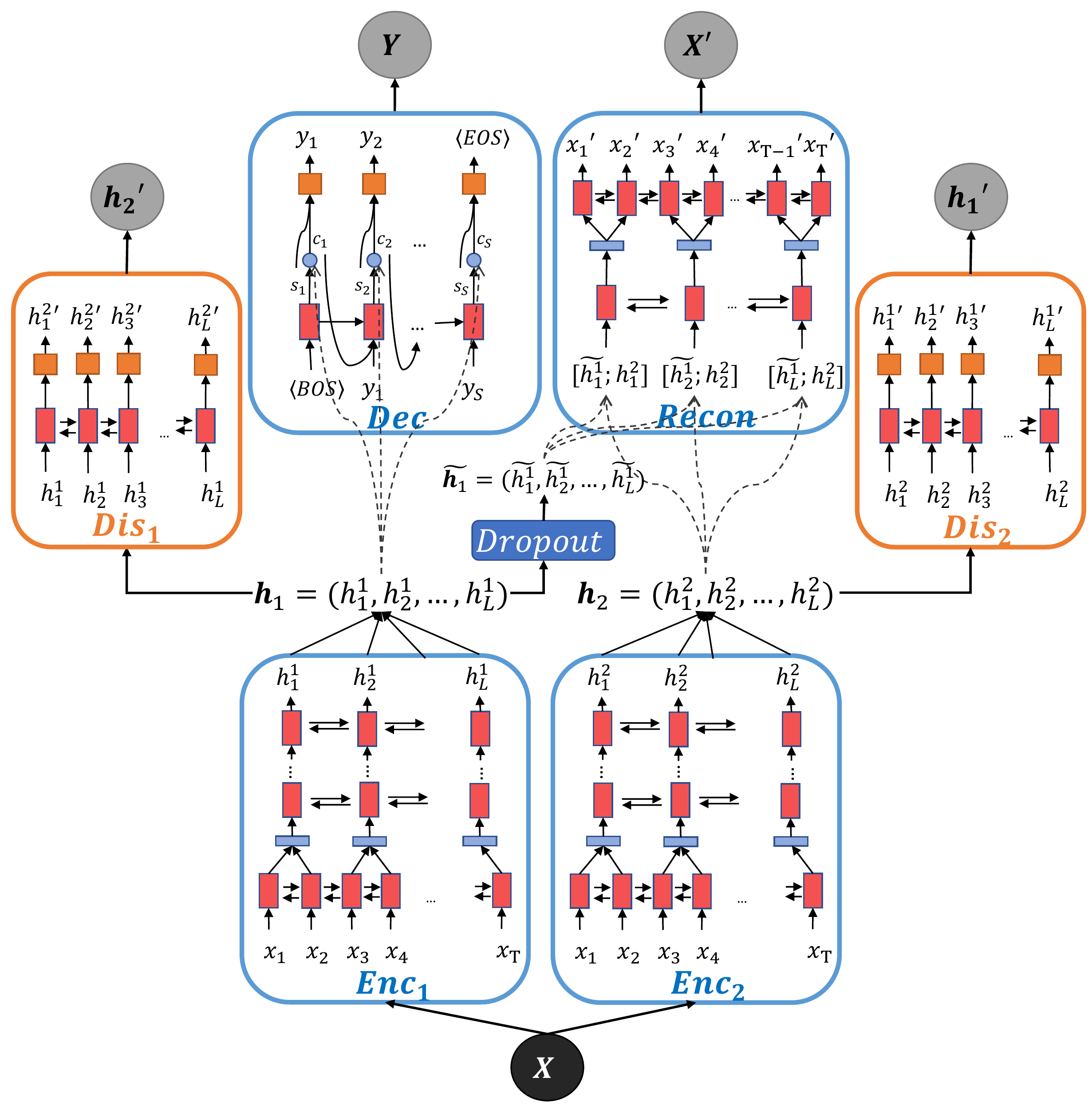}
    \caption{NIESR: The two encoders $Enc_1$ and $Enc_2$ are BLSTM-based feature extractors that encode the input sequence $\mathbf{x}$ into representations $\mathbf{h}^1$ and $\mathbf{h}^2$. The two encodings are disentangled by adversarially training the two disentanglers, $Dis_1$ and $Dis_2$, which aim to predict one embedding from another. $Dec$ is an attention-based decoder that generates the target $\mathbf{y}$ characters from $\mathbf{h}^1$. $Recon$ is a BLSTM-based reconstructor that decodes $\mathbf{h}^2$ and the noisy $\widetilde{\mathbf{h}}^1$ back to the input-sequence $\mathbf{x}$}
    \label{fig:Model}
\end{figure}

The NIESR model comprises five types of modules: (1) encoders $Enc_1$ and $Enc_2$ that map input data to the encodings $\mathbf{h}^1$ and $\mathbf{h}^2$, respectively, (2) a decoder $Dec$ that infers target $\mathbf{y}$ from $\mathbf{h}^1$, (3) a dropout layer that converts $\mathbf{h}^1$ into its noisy version $\widetilde{\mathbf{h}}^1$, (4) a reconstructor $Recon$ that reconstructs input data from $[\widetilde{\mathbf{h}}^1, \mathbf{h}^2]$, and (5) two adversarial disentanglers $Dis_1$ and $Dis_2$ that try to infer each embedding ($\mathbf{h}^1$ or $\mathbf{h}^2$) from the other. Figure~\ref{fig:Model} shows the complete NIESR model.

The encoder $Enc_1$ and decoder $Dec$ follow the base model design as described in Section~\ref{basemodel}, i.e., an attention-based Seq2Seq model for the speech recognition task. $Enc_2$ is designed to have exactly the same structure as $Enc_1$. The dropout layer is introduced to make $\widetilde{\mathbf{h}}^1$ an unreliable source of information for reconstruction, which influences the reconstruction task to extract all information about $\mathbf{x}$ into $\mathbf{h}^2$~\cite{jaiswal2018unsupervised}. $Recon$ is modeled as a stack of BLSTM layers interspersed with novel upsampling layers, which perform decompression by splitting information in each time-frame to two frames. This is the inverse of the subsampling layers~\cite{zhang2017very} used in $Enc_1$ and $Enc_2$. The upsampling operation is formulated as:
\begin{align}
    \label{upsample}
    &[u_{2i-1},u_{2i}]=\text{BLSTM}([\widetilde{h_i^1},h_i^2], s_{i-1}) \\
    &o_{2i}=Pu_{2i} \quad , \quad o_{2i-1}=Pu_{2i-1} 
\end{align}
where $[\cdot,\cdot]$ represents concatenation, $o$ is the output, and $P$ is a learned projection matrix.

The adversarial disentanglers $Dis_1$ and $Dis_2$ model the UAI disentanglement constraint discussed in Section~\ref{subsec:uai} following previous works~\cite{jaiswal2018unsupervised,jaiswal2019ropad,jaiswal2019unified}. $Dis_1$ tries to predict $\mathbf{h}^2$ from $\mathbf{h}^1$ and $Dis_2$ tries to do the inverse. This is directly opposite to the desired independence between $\mathbf{h}^1$ and $\mathbf{h}^2$. Thus, training $Dis_1$ and $Dis_2$ adversarially against the rest of the model helps achieve the independence goal. Unlike previous works~\cite{jaiswal2018unsupervised,jaiswal2019ropad,jaiswal2019unified}, the encodings $\mathbf{h}^1$ and $\mathbf{h}^2$ for this work are vector-sequences instead of single vectors: $\mathbf{h}^1=(h_1^1,h_2^1,\ldots,h_L^1)$ and $\mathbf{h}^2=(h_1^2,h_2^2,\ldots,h_L^2)$. Na\"ive instantiations of the disentanglers would perform frame-specific predictions of $h^2_i$ from $h^1_i$ and vice versa. However, each pair of $h^1_i$ and $h^2_i$ generated at the time-step $i$ contains information not only from frame $i$ but also from other frames across the time-span. This is because $Enc_1$ and $Enc_2$ are modeled as RNNs. Therefore, a better method to perform disentanglement for sequential representations is to use the whole series of $\mathbf{h}^1$ or $\mathbf{h}^2$ to estimate every element of the other. Hence, we model $Dis_1$ and $Dis_2$ as BLSTMs.

The proposed NIESR model is optimized by adopting the UAI training strategy~\cite{jaiswal2018unsupervised,jaiswal2019unified}, i.e., playing a game where we treat $Enc_1$, $Enc_2$, $Dec$, and $Recon$ as one player $\mathbf{P_1}$, and $Dis_1$ and $Dis_2$ as the other player $\mathbf{P_2}$. The model is trained using a scheduled update scheme where we freeze the weights of one player model when we update the weights of the other. The training objective comprises three tasks: (1) predicting transcriptions from the input signal, (2) reconstruction of the input, and (3) adversarial prediction of each of $\mathbf{h}^1$ and $\mathbf{h}^2$ from the other. The objective of the first task is written as Equation~\ref{L_y}. The goal for the reconstruction task is to minimize the mean squared error (MSE) between $\mathbf{x}$ and the reconstructed $\mathbf{x'}$:
\begin{align}
    L_x &= \ \text{MSE}(Recon([\psi(Enc_1(\mathbf{x})),Enc_2(\mathbf{x})]),\mathbf{x})
\end{align}
where $\psi$ means dropout. The training objective for the disentanglers is to minimize the MSE between embeddings predicted by the disentenglers and the embeddings generated from the encoder. However, that of the encoders is to generate $\mathbf{h}^1$ and $\mathbf{h}^2$ that are not predictive of each other. Hence, in the scheduled update scheme, the targets $\mathbf{t}^1$ and $\mathbf{t}^2$ for the disentanglers are different when updating the player models $P_1$ versus $P_2$, following~\cite{jaiswal2019unified}. The loss can be written as:
\begin{align}
    L_d &= \text{MSE}(Dis_1(Enc_1(\mathbf{x})), \mathbf{t}^1)\\
    & \ \ \ \ \ \ \ \ \ \ + \text{MSE}(Dis_2(Enc_2(\mathbf{x})),\mathbf{t}^2)) 
\end{align}
where $\mathbf{t}^1$ and $\mathbf{t}^2$ are set as $\mathbf{h}^2$ and $\mathbf{h}^1$, respectively, when updating $\mathbf{P_2}$ but are set to random vectors when updating $\mathbf{P_1}$.

Overall, the model is trained through backpropagation by optimizing the objective described in Equation~\ref{eq:obj}, where the loss-weights $\alpha$, $\beta$, and $\gamma$ are hyperparameters, which are decided by the performance on the development set.
\begin{equation}
    L = \alpha L_{y}+\beta L_x+\gamma L_d \label{eq:obj}
\end{equation}
Inference with NIESR involves a forward pass of data through $Enc_1$ followed by $Dec$. Hence, the usage and computational cost of NIESR for inference is the same as the base model.
\section{Experiments}
\label{sec:experiments}

The effectiveness of NIESR is quantified through the performance improvement achieved by adopting the invariant learning framework. We provide experimental results on speech recognition on three benchmark datasets: the Wall Street Journal Corpus (WSJ0)~\cite{paul1992design}, CHiME3~\cite{barker2015third}, and TIMIT~\cite{timit}. We additionally provide results on the combined WSJ0+CHiME3 dataset.

\subsection{Datasets}
\mypar{WSJ0:} This dataset is a collection of readings of the Wall Street Journal. It contains 7,138 utterances in the training set, 410 in the development set, and 330 in the test set. We use 40-dimensional log Mel filterbank features as the model input, and normalize the transcriptions to capitalized character sequences.

\mypar{CHiME3:} CHiME3 dataset contains: (1) WSJ0 sentences spoken in challenging noisy environments (real data) and (2) WSJ0 readings mixed with four different background noise (simulated data). The real speech data was recorded in five noisy environments using a six-channel tablet-based microphone array. Training data consists of 1,999 real noisy utterances from four speakers, and 7,138 simulated noisy utterances from 83 speakers in the WSJ0 training set. In total, there are 3,280 utterances in the development set, and 2,640 utterances in the test set containing both real and simulated data. The speakers in training, development, and test set are mutually different. In our experiments, we follow~\cite{meng2018speaker} to use far-field speech from the fifth microphone channel for all sets. We adopt the same input-output setting for CHiME3 as WSJ0.

\mypar{TIMIT:} This corpus contains a total of 6,300 sentences, with 10 sentences spoken by 630 speakers each with 8 different dialects. Among them, utterances from 168 different speakers are held-out as the test set. We further select sentences from 4 speakers of each dialect group, i.e., 32 speakers in total, from the remaining data to form the development set. Thus, all speakers in training, development, and test sets are different. Models were trained on 80 log Mel filterbank features and capitalized character sequences were treated as targets.

\subsection{Experiment Setup}

We train the base model without using invariance induction, i.e., the model consisting of $Enc$ and $Dec$ (Section~\ref{basemodel}), as a baseline. We feed the whole sequence of spectra features to $Enc$ and get the predicted character sequence from $Dec$. We use a stack of two BLSTMs with a subsampling layer (as described in Section~\ref{basemodel}) in between for $Enc$. $Dec$ is implemented as a single layer LSTM combined with attention modules introduced in Section~\ref{basemodel}. All the models were trained with early stopping with 30 epochs of patience and the best model is selected based on the performance on the development set. Other model and training hyperparameters are listed in Table~\ref{basehyper}.

\begin{table}[]
\centering
\caption{Hyperparameters for the base model.}
\setlength{\tabcolsep}{1.0em} 
\vspace{-5pt}
\begin{tabular}{|l|l|}
\hline
\textbf{Item}                     & \textbf{Setting} \\ \hline
$Enc$ and $Dec$ LSTM Dimension    & 200     \\ \hline
Subsampling Projected Dimension & 200     \\ \hline
Attention Dimension               & 200     \\ \hline
Attention Convolution Channel     & 10      \\ \hline
Attention Convolution Kernel Size & 100     \\ \hline
Optimizer                         & Adam    \\ \hline
Learning Rate                     & 5e-4    \\ \hline
\end{tabular}
\label{basehyper}
\vspace{-10pt}
\end{table}

\begin{table}[t]
\centering
\caption{Hyperparameters for the NIESR model.}
\setlength{\tabcolsep}{1.0em} 
\vspace{-5pt}
\begin{tabular}{|l|l|}
\hline
\textbf{Item}                     & \textbf{Setting} \\ \hline
$Recon$ LSTM Dimension    & 300     \\ \hline
Upsampling Projected Dimension & 200     \\ \hline
$Dis_1$, $Dis_2$ Dimension             & 200     \\ \hline
Dropout layer rate              & 0.4 \\ \hline
Optimizer                         & Adam    \\ \hline
Learning Rate for $\mathbf{P_1}$           & 5e-4    \\ \hline
Learning Rate for $\mathbf{P_2}$           & 1e-3   \\ \hline
$\alpha$, $\beta$, $\gamma$ for WSJ0 & 100, 10, 1 \\ \hline
$\alpha$, $\beta$, $\gamma$ for CHiME3 & 100, 1, 0.5 \\ \hline
$\alpha$, $\beta$, $\gamma$ for TIMIT & 100, 50, 1 \\ \hline
\end{tabular}
\label{modelhyper}
\end{table}

We augment the base model with $Enc_2$, $Recon$, $Dis_1$, and $Dis_2$, while treating $Enc$ as $Enc_1$, to form the NIESR model. $Enc_2$ has the same hyperparameter setting and structure as $Enc_1$. $Recon$ is modeled as a cascade of a BLSTM layer, an upsampling layer, and another BLSTM layer. $Dis_1$ and $Dis_2$ are implemented as BLSTMs followed by two fully-connected layers. We update the player models $\mathbf{P_1}$ and $\mathbf{P_2}$ in the frequency ratio of $1:5$ in our experiments. Hyperparameters for $Enc_1$ and $Dec$ are the same as the base model. Additional hyperparameters for NIESR are summarized in Table~\ref{modelhyper}.

We further provide results of a stronger baseline model that utilizes \emph{labeled} nuisances $z$ (speakers for WSJ0, speakers and noise environment condition for CHiME3, speakers and dialect groups for TIMIT) with the gradient reversal layer (GRL)~\cite{ganin2014unsupervised} to learn invariant representations. Specifically, the model consists of $Enc$, $Dec$, and a classifier with a GRL between the embedding learned from $Enc$ and the classifier, following the standard setup in~\cite{ganin2014unsupervised}. The target for the classifier is to predict $z$ from the embedding while the direction of the training gradient to $Enc$ is flipped. We denote this model as \textbf{Spk-Inv} for speaker-invariance, \textbf{Env-Inv} for environment-invariance in CHiME3, and \textbf{Dial-Inv} for dialect-invariance in TIMIT.

\begin{table}[]
\centering
\caption{Speech recognition performance as CER (\%). Values in parentheses show relative improvement (\%) over Base model.}
\setlength{\tabcolsep}{0.9em} 
\vspace{-5pt}
\begin{tabular}{|c|c|c|c|}
\hline
\textbf{Model} & \textbf{WSJ0} & \textbf{CHiME3} & \textbf{TIMIT} \\ \hline
Base & 12.95 & 44.61 & 28.76 \\ \hline
Spk-Inv & 12.31 (4.94) & 43.93 (1.52) & 28.45 (1.08) \\ \hline
Env-Inv & -- & 42.61 (4.48) & -- \\ \hline
Dial-Inv & -- & -- & 28.29 (1.63) \\ \hline
NIESR & \textbf{12.24 (5.48)} & \textbf{41.86 (6.16)} & \textbf{26.86 (6.61)} \\ \hline
\end{tabular}
\label{basicresults}
\end{table}

\subsection{ASR Performance on Benchmark Datasets}

Table~\ref{basicresults} summarizes the results at end-to-end ASR on WSJ0, CHiME3, and TIMIT datasets. Results show that NIESR achieves 5.48\%, 6.16\%, and 6.61\% relative improvements over base model on WSJ0, CHiME3, and TIMIT, respectively, and demonstrates the best CER among all methods.

\subsection{Invariance to Nuisance Factors}

In order to examine whether a latent embedding is invariant to nuisance factors $z$, we calculate the accuracy of predicting the factor $z$ from the encoding. Specifically, this is calculated by training classification networks (BLSTM followed by two fully-connected layers) to predict $z$ from the generated embeddings. Table~\ref{nuisance} presents results of this experiment, showing that the $\mathbf{h}^1$ embedding of the NIESR model, which is used for ASR, contains less nuisance information than the $\mathbf{h}$ encoding of the base, Spk-Inv, and Env-Inv models. In contrast, the $\mathbf{h}^2$ embedding of NIESR contains most of the nuisance information, showing that nuisance factors migrate to this embedding, as expected.

\begin{table}[t]
\centering
\caption{Results of predicting nuisance factor $z$ from  learned representations as accuracy. Env stands for environment.}
\setlength{\tabcolsep}{0.85em} 
\vspace{-5pt}
\begin{tabular}{|c|c|c|c|}
\hline
\multirow{2}{*}{\textbf{Dataset}} & \multirow{2}{*}{\textbf{Predict $z$ from}} & \multicolumn{2}{c|}{\textbf{Accuracy}} \\ \cline{3-4} 
 &  & \textbf{$z$ : Speaker} & \textbf{$z$ : Env} \\ \hline
\multirow{4}{*}{WSJ0} & $\mathbf{h}$ in Base Model & 67.91 & -- \\ \cline{2-4} 
 & $\mathbf{h}$ in Spk-Inv & 65.60 & -- \\ \cline{2-4} 
 & $\mathbf{h}_1$ in NIESR & \textbf{63.35} & -- \\ \cline{2-4} 
 & $\mathbf{h}_2$ in NIESR & 97.92 & -- \\ \hline
\multirow{5}{*}{CHiME3} & $\mathbf{h}$ in Base Model & 38.52 & 69.24 \\ \cline{2-4} 
 & $\mathbf{h}$ in Spk-Inv & 37.91 & 69.11 \\ \cline{2-4} 
 & $\mathbf{h}$ in Env-Inv & 38.84 & 66.44 \\ \cline{2-4} 
 & $\mathbf{h}_1$ in NIESR & \textbf{35.87} & \textbf{63.45} \\ \cline{2-4} 
 & $\mathbf{h}_2$ in NIESR & 92.28 & 97.05 \\ \hline
\end{tabular}
\label{nuisance}
\end{table}

\subsection{Additional Robustness through Data Augmentation}

Training with additional data that reflects multiple variations of nuisance factors helps models generalize better. In this experiment, we treat the CHiME3 dataset, which contains WSJ0 recordings with four different types of noise, as a noisy augmentation for WSJ0. We train the base model and NIESR on the augmented dataset, i.e. WSJ0+CHiME3, and test on the original CHiME3 and WSJ0 test sets separately. Table~\ref{augmentresult} summarizes the results on this experiment, showing that training with data augmentation provides improvements on both CHiME3 and WSJ0 datasets compared to the results in Table~\ref{basicresults}. It is important to note that the NIESR model trained on the augmented dataset achieves 14.44\% relative improvement on WSJ0 as compared to the base model trained on the same. This is because data augmentation provides additional information about potential nuisance factors to the NIESR model and, consequently, helps it ignore these factors for the ASR task, even though pairwise data is not provided to the model like~\cite{liang2018learning}. Hence, results show that the NIESR model can be easily combined with data augmentation to further enhance the robustness and nuisance-invariance of the learned features.

\begin{table}[]
\centering
\caption{Test results of models trained on the WSJ0+CHiME3 augmented dataset as CER (\%). Values in parentheses show the relative improvement (\%) over Base model.}
\setlength{\tabcolsep}{2.0em}
\vspace{-5pt}
\begin{tabular}{|c|c|c|}
\hline
\textbf{Model} & \textbf{WSJ0} & \textbf{CHiME3} \\ \hline
Base & 9.35 & 41.55 \\ \hline
Spk-Inv & 8.62 (7.81) & 40.77 (1.88) \\ \hline
Env-Inv & 9.17 (1.93) & 40.27 (3.08) \\ \hline
NIESR & \textbf{8.00 (14.44)} & \textbf{38.35 (7.7)} \\ \hline
\end{tabular}
\label{augmentresult}
\vspace{-8pt}
\end{table}
\section{Conclusion}
\label{sec:conclusion}
We presented NIESR, an end-to-end speech recognition model that adopts the unsupervised adversarial invariance framework for invariance to nuisances without requiring any knowledge of potential nuisance factors. The model works by learning a split representation of data through competition between the recognition and an auxiliary data reconstruction task. Results of experimental evaluation demonstrate that the proposed model achieves significant boosts in performance on ASR.
\section{Acknowledgements}
\label{sec:acknowledgements}
This material is based on research sponsored by DARPA under agreement number FA8750-18-2-0014. The U.S. Government is authorized to reproduce and distribute reprints for Governmental purposes notwithstanding any copyright notation thereon. The views and conclusions contained herein are those of the authors and should not be interpreted as necessarily representing the official policies or endorsements, either expressed or implied, of DARPA or the U.S. Government.

\bibliographystyle{IEEEtran}
\bibliography{main}

\end{document}